\documentclass[conference]{IEEEtran}

\IEEEoverridecommandlockouts

	\usepackage[utf8]{inputenc} 
	\usepackage[T1]{fontenc}

\ifCLASSINFOpdf
  \usepackage[pdftex]{graphicx}
  
\else
 
\fi

\usepackage[cmex10]{amsmath}
\usepackage[font={footnotesize}]{caption}

\usepackage{multirow}
\usepackage{array}
\usepackage[lofdepth,lotdepth,font=footnotesize]{subfig}

\usepackage{adjustbox}

\usepackage{xfrac} 
\usepackage{multirow} 
\usepackage{graphicx}
\usepackage{grffile}
\usepackage[style=ieee]{biblatex}
\AtBeginDocument{%
  \renewcommand\tablename{TABLE}
}

\AtBeginDocument{%
  \renewcommand\abstractname{Abstract}
}

\begin{document}

%
\title{Sıralama Sorunu Olarak Nispi Derinlik Tahmini\\
Relative Depth Estimation as a Ranking Problem}

\author{\IEEEauthorblockN{Alican Mertan}
\IEEEauthorblockA{Department of Computer Engineering\\
Istanbul Technical University\\ Istanbul, Turkey \\
Email: mertana@itu.edu.tr}
\and
\IEEEauthorblockN{Damien Jade Duff}
\IEEEauthorblockA{Department of Computer Engineering\\
Istanbul Technical University\\ Istanbul, Turkey \\
Email: djduff@itu.edu.tr}
\and
\IEEEauthorblockN{Gözde Ünal}
\IEEEauthorblockA{Department of Computer Engineering\\
Istanbul Technical University\\ Istanbul, Turkey \\
Email: gozde.unal@itu.edu.tr}
}

\IEEEoverridecommandlockouts
\IEEEpubid{\makebox[\columnwidth]{978-1-7281-7206-4/20/\$31.00~\copyright2020 IEEE \hfill} \hspace{\columnsep}\makebox[\columnwidth]{ }}

\maketitle

\begin{ozet}
Bu bildiride, tekil görüntüden nispi derinlik tahmin etme problemini, sıralama problemi olarak ele alan ve çözüm sunan yeni bir yöntem sunuyoruz. Derinlik tahmini problemini bu şekilde yeniden formüle ederek, sıralama problemi üzerine var olan literatürden faydalanabildik ve mevcut bilgileri daha iyi sonuçlar elde etmek için uygulayabildik. Bu amaçla, sıralama literatüründen ödünç aldığımız ağırlıklandırılmış ListMLE kayıp fonksiyonunu nispi derinlik tahmini problemine kazandırdık. Ayrıca, yöntemimizin daha güçlü olduğu piksel derinliği sıralama doğruluğunu dikkate alan yeni bir metrik getirdik.
\end{ozet}
\begin{IEEEanahtar}
derin öğrenme, derinlik tahmini, nispi derinlik tahmini, sıralamayı öğrenme.
\end{IEEEanahtar}

\begin{abstract}
We present a formulation of the relative depth estimation from a single image problem, as a ranking problem. By reformulating the problem this way, we were able to utilize literature on the ranking problem, and apply the existing knowledge to achieve better results. To this end, we have introduced a listwise ranking loss borrowed from ranking literature, weighted ListMLE, to the relative depth estimation problem. We have also brought a new metric which considers pixel depth ranking accuracy, on which our method is stronger.

\end{abstract}

\IEEEpeerreviewmaketitle

\IEEEpubidadjcol

\section{Introduction}

Depth is one of the key elements of a scene. Estimation of depth is a pathway to understanding the 3D structure of a scene, which can be useful in a wide variety of tasks such as object detection or semantic segmentation \cite{cao_estimating_2017}, robotic tasks such as obstacle avoidance \cite{michels_high_2005}, or depth-aware image processing tasks such as virtual or augmented reality technologies \cite{li_two-streamed_2017}. 

In particular, recent research has focused on the problem of estimating the depth from a single image, even though there are other alternatives that one can use to estimate the depth e.g. binocular depth estimation, structure from motion, or simply usage of depth sensors. While all of these alternatives can be used successfully, each have their own limitations. For binocular depth estimation, two cameras are needed and they need to be carefully calibrated \cite{mal_sparse--dense:_2018}. Even if this is given, it is still an open area of research since corresponding points in two stereo images need to be found. Moreover, it does not work well on featureless regions \cite{mal_sparse--dense:_2018} and can have a limited range \cite{michels_high_2005}. Another option is to incorporate motion to estimate the depth. This is referred to as structure from motion (SfM) and requires multiple images collected with a changing viewpoint. As with binocular depth estimation, it is not good on featureless regions in the image and most methods assume a static scene, an assumption which does not always hold \cite{ranjan_competitive_2018}. Finally, different kinds of sensors exist to measure depth. However, the quality of this estimation depends on the  quality of the hardware which can be very costly and power consuming, have a short range, or produce sparse depth maps since they don't work on all surfaces, or are sensitive to light \cite{mal_sparse--dense:_2018, garg_unsupervised_2016}. As we can see, among these options depth estimation from a single image is potentially the most robust one with minimum requirements and limitations.

However, estimating depth from a single image is a strictly harder problem. As was observed at least as early as Helmholtz, the same 2D image can be produced by an endless number of 3D scenes, meaning that there is a one to many relation which creates an ill-posed problem \cite{li_two-streamed_2017, laina_deeper_2016, eigen_depth_2014}. Yet, in reality, many of these 3D scenes are not plausible or probable in our world and humans can estimate depth from a single image using prior knowledge. Therefore statistical learning methods can be utilized to solve the problem.

In recent years, deep learning has been used to estimate depth from a single image with great success. Eigen, Puhrsch, and Fergus \cite{eigen_depth_2014} estimated depth with a two networks where first network makes a coarse prediction with global cues and the second one refines the prediction of the first one using local cues. This group also implemented a scale invariant loss to overcome scale ambiguity. Li, Klein, and Yao \cite{li_two-streamed_2017} tackled the problem with a multitask framework where a spatial depth gradient is estimated as well as depth in order to increase the quality of the details. Cao, Wu, and Shen \cite{cao_estimating_2017} approached the problem as a classification problem by estimating the depth range of a pixel instead of regressing to the exact depth value which allow them to have a confidence value for their results. Laina et al. \cite{laina_deeper_2016} employed ResNet with novel upsampling blocks to increase the output resolution, and implemented a new loss which puts more emphasis on small residuals. Fu et al. \cite{fu_deep_2018} reformulated the problem as an ordinal regression problem and achieved higher accuracy with faster convergence.

While the aforementioned works achieved great success, they have only worked on images coming from similar distributions as their training data sets, for example, if trained on outdoor images they do not work well on indoor images. Neither do they generalise well under changes to camera parameters. In order to be able to estimate depth in an unconstrained environment, diverse data sets are needed. However collecting a diverse data set with absolute depth annotation, i.e. actual depth in meters, is infeasible \cite{chen_single-image_2016}. It might be expected that training from multiple absolute depth datasets can provide us with the desired generalisation but in practice this is not the case. One hypothesis for why this is the case is that existing absolute depth datasets do not provide an even distribution over images to be found in the wild; another that absolute depth is an inherently more difficult problem and harder to learn.

Regardless, to be able to estimate depth in an unconstrained environment, data sets with relative depth annotations, i.e. ordinal relations of the pixels, are collected \cite{chen_single-image_2016, xian_monocular_2018, li_megadepth:_2018, chen_learning_2019}. The advantage of data sets with relative depth annotations is the fact that they consist of diverse set of images collected from unconstrained environments and they allow better generalization in the wild.

In this work, we have trained a neural network in a supervised manner to estimate relative depth from a given single image. Our contribution is two-fold:
\begin{itemize}
    \item We have formulated the relative depth estimation problem as a ranking problem and applied a listwise ranking loss to learn a depth estimator.
    \item We have introduced mean average precision (MAP) as a metric for evaluation for the relative depth estimation problem.
\end{itemize}{}

\section{Related Work}

\textbf{Relative Depth Estimation} Chen et al. \cite{chen_single-image_2016} introduced a new data set called depth in the wild (DIW) that consists of RGB images and the ordinal relation of only one pixel pair per image as a ground truth annotation. They also devised the loss shown in eq. \ref{eq:pairwise}, which will be referred as pairwise loss, for learning a relative depth estimator from pairwise ordinal relations.
\begin{equation} \label{eq:pairwise}
    L_k=\left\{\begin{array}{ll}{\log \left(1+\exp \left(-z_{i k}+z_{j k}\right)\right),} & {r_{k}=+1} \\ {\log \left(1+\exp \left(z_{i_{k}}-z_{j_{k}}\right)\right),} & {r_{k}=-1} \\ {\left(z_{i_{k}}-z_{j_{k}}\right)^{2},} & {r_{k}=0}\end{array}\right.
\end{equation}
Here, $z_{i_{k}}$ is the depth prediction of the network for the pixel $i$ in the $k$th image and the $z_{j_{k}}$ is the depth prediction of the network for the pixel $j$ in the $k$th image. $r_k$ shows the relative depth relation between the two pixels in the $k$\textsuperscript{th} image where $r_k=+1$ means pixel $i$ is closer, $r_k=-1$ means pixel $j$ is closer and $r_k=0$ means the two pixels are at the same depth. This loss is summed across annotated pairs. Note that there is one pixel pair annotated per image.

For $r_k=0$, the loss penalizes the network for outputting different depth values for the two pixels (note that the actual value that network outputs is irrelevant; this loss function just pushes network to output same value for the two pixel if they are at the same depth, with the degree of push dependent only how different the two estimated depths are). For $r_k = +1$ and $r_k = -1$, the loss function pushes the network to increase the distance between the estimated depth for the given pixels in the correct way depending on the relative depth relation.

Several other data sets with relative depth annotations have been introduced \cite{li_megadepth:_2018, xian_monocular_2018, chen_learning_2019}. They have used the same pairwise loss to learn from ordinal relations.

\textbf{Learning to Rank} Ranking methods mainly fall into two categories: pairwise methods and listwise methods. During the learning process, pairwise methods use a loss function that considers pairs of items for a given query (such as the pairwise loss in eq. \ref{eq:pairwise} that has been used in the depth estimation in the wild research so far). On the other hand, listwise methods define losses over the set of permutations of the items. A permutation here means an ordered set of the items, e.g. for three items $a$, $b$, and $c$, $(a,b,c)$ is a permutation which ranks $a$ highest and $c$ lowest. Listwise methods perform better in the ranking literature compared to pairwise methods since they take into account more information: complete ranking \cite{cao_learning_2007}.

Listwise methods started with Cao et al. \cite{cao_learning_2007}, who  defined a probability called "top-one probability" for items in the permutation:
\begin{equation}
P_{s}(j)=\frac{exp\left(s_{j}\right)}{\sum_{k=1}^{n} exp\left(s_{k}\right)}
\end{equation}
Here, $s_j$ is the score of item $j$. Using cross entropy to measure the distance between ground truth top-one probabilities and predicted top-one probabilities, the proposed loss takes the form:
\begin{equation}\label{eq:listwise}
    L\left(y, z\right)=-\sum_{j=1}^{n} P_{y}(j) \log \left(P_{z}(j)\right)
\end{equation}
Here, $y$ is the ground truth permutation and $z$ is the predicted permutation. The problem with this approach is the need for ground truth scores which can be missing in data sets.

Instead of calculating the distance between probability distributions, it is shown that one can try to minimize the negative log likelihood of permutation probability defined by Xia et al. \cite{xia_listwise_2008} as:

\begin{equation}\label{eq:listMLE}
\begin{aligned}
    P(y | x ; g) =\prod_{i=1}^{n} \frac{\exp (g(x_{y(i)}))}{\sum_{k=i}^{n} \exp (g(x_{y(k)}))}\\
    L(y, z) =-\log P(y | x ; g) 
\end{aligned}
\end{equation}

Here, $x_{y(i)}$ is the item that is ranked in the $i$\textsuperscript{th} position in the ground truth permutation $y$, and $g()$ is the ranking function. Note that the probability will take higher values when the predicted scores are consistent with the ground truth permutation. It is also proven that the above ranking loss (named ListMLE) is consistent, sound, has nice mathematical properties such as being continuous, differentiable, and convex, and can be calculated in linear time, which makes it theoretically better compared than the loss given in eq. \ref{eq:listwise} \cite{xia_listwise_2008}.

Two improvements have been made to eq. \ref{eq:listMLE}, by proposing weighting for the loss. Lan et al. \cite{lan_position-aware_2014} showed that weighting with a decreasing function makes the loss capture position importance well. Chen et al. \cite{chen_ranking_2009} showed that weighting with gain and discount functions, used in ranking-measure metric normalized discounted cumulative gain (NDCG), makes the loss a tighter upper bound for NDCG.

In this work, we propose to introduce the ListMLE loss into an autoencoder model as a solution to the relative depth estimation problem. To the best of our knowledge, listwise ranking losses were not used in depth estimation before.

\section{Proposed Method}

\subsection{Relative Depth Estimation as Ranking Problem}

A ranking problem is defined as information retrieval problem where there is a set of queries $ Q = \{q^1, ... , q^m\}$, for each query $q^i$ there is set of items $I^i = \{ i^i_1, ... ,i^i_n  \} $ to be ranked based on their relevance to the given query and ground truth ranking $R^i = \{ r^i_1, ..., r^i_n \} $ where $r^i_1$ is the rank of item with index $i$. Alternatively, one can have list of ground truth scores of each item $S^i = \{ s^i_1, ..., s^i_n \}$ which will give the complete ground truth ranking when sorted in descending order. The aim is to find a ranking function $g()$ which produces scores $g(q^i,i^i_a) = z^i_a$ such as $z^i_a > z^i_b$ $\forall \; a,\,b \; s.t. \; r^i_a < r^i_b$. 

The depth estimation problem can be reformulated as a ranking problem where there is a set of input RGB images $X = \{ x^1, ..., x^m \}$, for each $x^i$ there is set of pixels ${Px}^i = \{ {px}^i_1, ..., {px}^i_n \}$ to be ranked based on their depth where $n = height \times width$ and there is a ground truth depth ranking $R^i = \{ r^i_1, ..., r^i_n \}$ showing the pixels' ordinal ranking. The aim is to find a ranking function $g()$, modeled by a neural network, which produces depth scores for each pixel $g(x^i) = Z $ where $ Z = \{ z^i_1, ..., z^i_n \}$. Learning can be done by minimizing the loss function calculating the loss using predicted and ground truth rankings.

\subsection{Listwise Ranking Loss}

ListMLE loss is used as a loss function in the learning process. The loss in the eq. \ref{eq:listMLE} can be rewritten as follows:

\begin{equation}\label{eq:listMLE_open}
    L(y,z)=\sum_{i=1}^{n-1}\left(-g\left(x_{y(i)}\right)+\ln \left(\sum_{s=i}^{n} \exp (g(x_{y(s)}))\right)\right)
\end{equation}{}

Intuitively, what it does is to increase the difference between scores of an item and the items ranked below it. The permutation probability that we try to maximize here is the Plackett-Luce model where the probability distribution is decomposed into stepwise conditional probabilities and for each conditional probability it is given that the items before it are ranked correctly \cite{lan_position-aware_2014, xia_listwise_2008}.

Since the data set that we have used has ground truth scores, we have used weighted ListMLE where the loss is weighted with gain and discount functions as described in \cite{chen_ranking_2009}:

\begin{equation}\label{eq:weighted_listMLE}
\begin{gathered}
    L(y,z)=\\
    \sum_{i=1}^{n-1} G(s_{y(i)}) D(i) \left(-g(x_{y(i)})+\ln \left(\sum_{s=i}^{n} \exp (g(x_{y(s)}))\right)\right)\\
    G(s) = 2^s-1\\
    D(s) = \frac{1}{\log (s+1)}
\end{gathered}
\end{equation}{}

Here, $y(i)$ returns the index of the item ranked i\textsuperscript{th} in the ground truth ranking, hence $s_{y(i)}$ is the score of the item ranked i\textsuperscript{th} in the ground truth ranking.

\subsection{Network Architecture}

In order to be able to compare our results and test the effectiveness of our proposed method, we have used the same network architecture as Xian et al. \cite{xian_monocular_2018} which will be referred to as EncDecResNet. The architecture is built upon a pre-trained ResNet. The final pooling layer, fully connected layer and softmax layer are removed and multi-scale feature fusion modules are added to the end to increase the resolution of the output. Please refer to the work of Xian et al. \cite{xian_monocular_2018} for more details.

\section{Experiments}

We have experimented on the ReDWeB data set. To the best of our knowledge, the only work that has been is done on the ReDWeB data set is the work of Xian et al. \cite{xian_monocular_2018}. We will be comparing our results with theirs.

Since the authors of \cite{xian_monocular_2018} didn't share their network weights, we have implemented and tested their system as described in their work. They have randomly sampled 3000 pixel pairs per image and calculated the loss for each pair by using the pairwise loss function. The results shared by the authors of \cite{xian_monocular_2018}, the results of our replication, and results of another replication by \cite{chen_learning_2019} can be seen in Table \ref{tab:redweb_diw}. Weighted human disagreement rate (WHDR) shows the percentage of incorrectly ordered points, therefore lower is better.

\begin{table}[h]
\centering
\caption{Results of training on ReDWeB data set, testing on DIW test split.}
\begin{adjustbox}{width=.6\columnwidth}
\begin{tabular}{|ll|}
\hline
\multicolumn{1}{|l}{Replications of EncDecResNet \cite{xian_monocular_2018}} & \multicolumn{1}{r|}{WHDR} \\ \hline
\multicolumn{1}{|l}{Our implementation}              & 16.28\%          \\ 
\multicolumn{1}{|l}{Implementation of Chen et al. \cite{chen_learning_2019}}          & 16.31\%                   \\ \hline
\multicolumn{1}{|l}{Original results from Xian et al. \cite{xian_monocular_2018}} & 14.33\%                   \\ \hline
\end{tabular}
\end{adjustbox}
\label{tab:redweb_diw}
\end{table}

We have trained the same architecture with our proposed weighted ListMLE loss on the ReDWeB data set and tested on the DIW test split to compare our results. For each input image, we sampled 500 points randomly and created the ground truth permutation. The results in the Table \ref{tab:pairwise_listwise} shows that we have achieved comparable results with our proposed loss. Note that we get similar performances by only using 500 points while they were using 3000 pixel pairs per image.

\begin{table}[h]
\centering
\caption{Comparison on DIW test split with different losses.}
\begin{adjustbox}{width=.6\columnwidth}
\begin{tabular}{|l|l|l|}
\hline
Network       & Loss             & \multicolumn{1}{r|}{WHDR} \\ \hline
EncDecResNet & pairwise         & \textbf{\%16.28}          \\ \hline
EncDecResNet & weighted ListMLE & \%16.33                   \\ \hline
\end{tabular}
\end{adjustbox}
\label{tab:pairwise_listwise}
\end{table}

In addition to the pairwise metric WHDR, we have also used average precision (AP) to evaluate the performance since AP is widely used in the literature to measure the performance of a ranking. It is formulated as:
\begin{equation}
    \label{eq:ap}
    \mathrm{AP}=\sum_{i=1}^{N} \operatorname{Precision}(i) ( \operatorname{Recall}(i) - \operatorname{Recall}(i-1) )
\end{equation}{}

Average precision is defined for binary labeled items. Therefore we have defined $R_k$ to be the ground truth ranking where items up to the  k\textsuperscript{th} point are labeled positively and our full metric then can be calculated as follows:
\begin{equation}
\begin{aligned}
    \label{eq:map}
    \mathrm{MAP} = \frac{1}{M} \sum_{i=1}^M \left( \frac{1}{N} \sum_{j=1}^{N} \operatorname{AP}(R_j^i) \right)
\end{aligned}
\end{equation}{}

Because the DIW data set has a ground truth ordinal depth annotation for only one pair of pixels per image, we have tested on the YouTube3D data set \cite{chen_learning_2019} where more the ground truth annotations are available per image. Results in the table \ref{tab:map_results} shows that our proposed method achieves better results with MAP as a metric. We conjecture that this is due to MAP penalizing more for mistakes done for the items ranked higher compared to mistakes done for items ranked lower whereas WHDR does not take this into account. 

\begin{table}[h]
\centering
\caption{Results of training on ReDWeB data set and testing on YouTube3D data set. Lower is better for WHDR and higher is better for MAP.}
\begin{adjustbox}{width=.6\columnwidth}
\begin{tabular}{|l|l|l|}
\hline
Loss & pairwise         & \multicolumn{1}{r|}{weighted listMLE} \\ \hline
WHDR & \textbf{\%27.19} & \%27.60                               \\ \hline
MAP  & \%70.36          & \textbf{\%70.67}                      \\ \hline
\end{tabular}
\end{adjustbox}
\label{tab:map_results}
\end{table}

\begin{figure}[h]
\centering
\subfloat{\includegraphics[width=0.85in,height=0.85in]{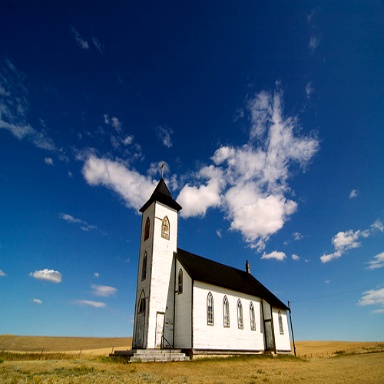}}
\subfloat{\includegraphics[width=0.85in,height=0.85in]{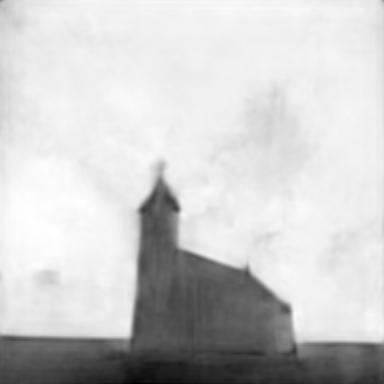}}  \subfloat{\includegraphics[width=0.85in,height=0.85in]{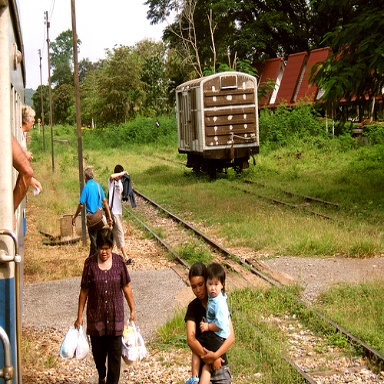}}
\subfloat{\includegraphics[width=0.85in,height=0.85in]{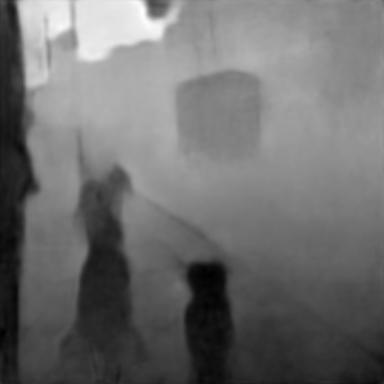}}\\[-2.2ex]
\subfloat{\includegraphics[width=0.85in,height=0.85in]{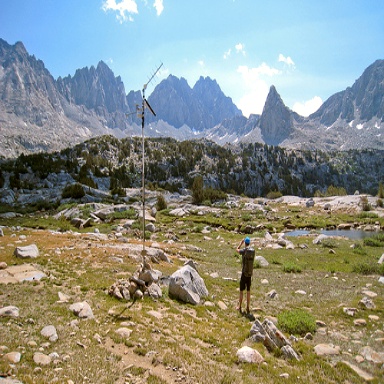}}
\subfloat{\includegraphics[width=0.85in,height=0.85in]{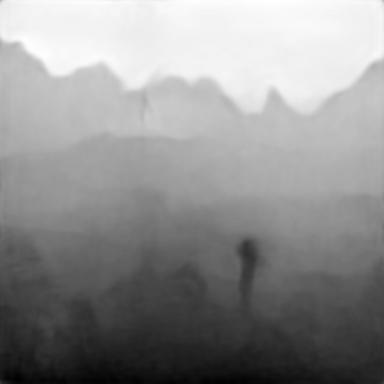}}
\subfloat{\includegraphics[width=0.85in,height=0.85in]{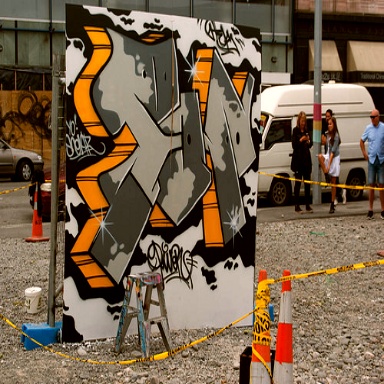}}
\subfloat{\includegraphics[width=0.85in,height=0.85in]{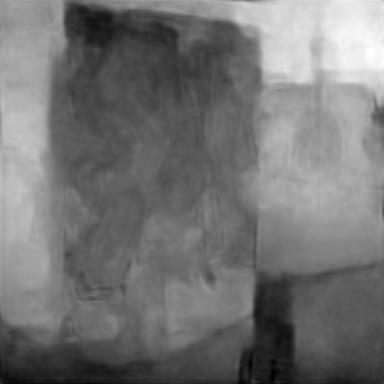}}\\[-2.2ex]
\subfloat{\includegraphics[width=0.85in,height=0.85in]{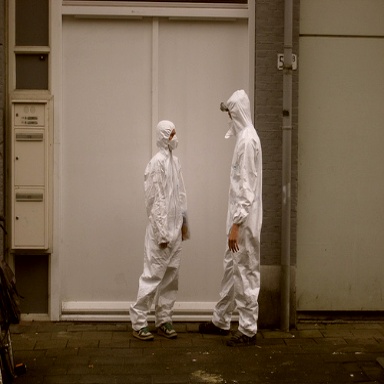}}
\subfloat{\includegraphics[width=0.85in,height=0.85in]{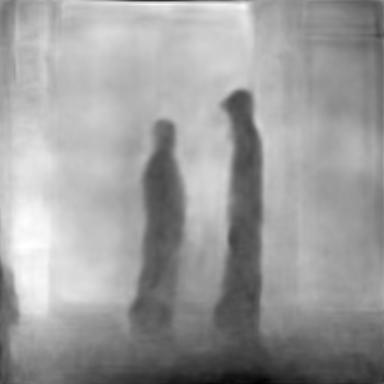}}
\subfloat{\includegraphics[width=0.85in,height=0.85in]{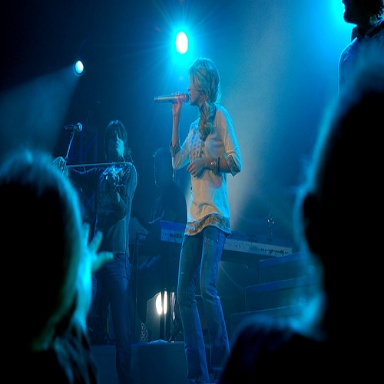}}
\subfloat{\includegraphics[width=0.85in,height=0.85in]{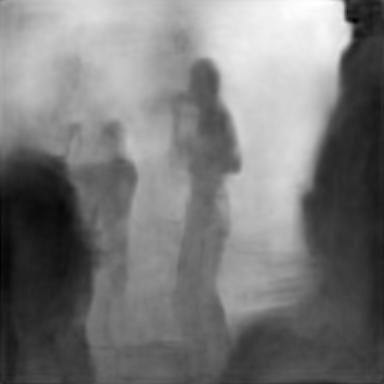}}
\caption{Predictions of our proposed system.}
\label{fig:results}
\end{figure}

Fig. \ref{fig:results} shows the prediction of our proposed system. It can be seen that the network captures the overall structure of the scene rather well. However, surfaces with distinct features propose a challenge for the system as a lot of global cues and prior knowledge need to be utilised to estimate the depth of that kind of surface correctly. An example can be seen in middle row, right column. Similarly, depth estimation for featureless patches also requires utilization of the global information. An example can be seen in bottom row, left column.

\section{Conclusion}

We have tackled the problem of estimating relative depth from a single image. Our contribution is modelling the problem as a ranking problem and applying a listwise ranking loss. We were able to achieve comparable results with our proposed approach. We have also introduced a new metric to evaluate the performance of the systems that accounts for the order of depths in an image, on which our method performs better.

\section*{Acknowledgement}

The work is supported by the Scientific and Technological
Research Council of Turkey (TÜBITAK), project 116E167.

\renewcommand*{\bibfont}{\footnotesize}
\printbibliography

@inproceedings{eigen_depth_2014,
	title = {Depth map prediction from a single image using a multi-scale deep network},
	booktitle = {Advances in neural information processing systems},
	author = {Eigen, David and Puhrsch, Christian and Fergus, Rob},
	year = {2014},
	pages = {2366--2374}
}

@inproceedings{li_two-streamed_2017,
	title = {A two-streamed network for estimating fine-scaled depth maps from single rgb images},
	booktitle = {Proceedings of the {IEEE} {International} {Conference} on {Computer} {Vision}},
	author = {Li, Jun and Klein, Reinhard and Yao, Angela},
	year = {2017},
	pages = {3372--3380}
}

@article{ranjan_competitive_2018,
	title = {Competitive {Collaboration}: {Joint} {Unsupervised} {Learning} of {Depth}, {Camera} {Motion}, {Optical} {Flow} and {Motion} {Segmentation}},
	journal = {arXiv preprint arXiv:1805.09806},
	author = {Ranjan, Anurag and Jampani, Varun and Balles, Lukas and Kim, Kihwan and Sun, Deqing and Wulff, Jonas and Black, Michael J.},
	year = {2018}
}

@inproceedings{chen_single-image_2016,
	title = {Single-image depth perception in the wild},
	booktitle = {Advances in {Neural} {Information} {Processing} {Systems}},
	author = {Chen, Weifeng and Fu, Zhao and Yang, Dawei and Deng, Jia},
	year = {2016},
	pages = {730--738}
}

@inproceedings{fu_deep_2018,
	title = {Deep ordinal regression network for monocular depth estimation},
	booktitle = {Proceedings of the {IEEE} {Conference} on {Computer} {Vision} and {Pattern} {Recognition}},
	author = {Fu, Huan and Gong, Mingming and Wang, Chaohui and Batmanghelich, Kayhan and Tao, Dacheng},
	year = {2018},
	pages = {2002--2011}
}

@inproceedings{michels_high_2005,
	title = {High speed obstacle avoidance using monocular vision and reinforcement learning},
	isbn = {1-59593-180-5},
	booktitle = {Proceedings of the 22nd international conference on {Machine} learning},
	publisher = {ACM},
	author = {Michels, Jeff and Saxena, Ashutosh and Ng, Andrew Y.},
	year = {2005},
	pages = {593--600}
}

@inproceedings{garg_unsupervised_2016,
	title = {Unsupervised cnn for single view depth estimation: {Geometry} to the rescue},
	booktitle = {European {Conference} on {Computer} {Vision}},
	publisher = {Springer},
	author = {Garg, Ravi and BG, Vijay Kumar and Carneiro, Gustavo and Reid, Ian},
	year = {2016},
	pages = {740--756}
}

@inproceedings{mal_sparse--dense:_2018,
	title = {Sparse-to-dense: {Depth} prediction from sparse depth samples and a single image},
	isbn = {1-5386-3081-8},
	booktitle = {2018 {IEEE} {International} {Conference} on {Robotics} and {Automation} ({ICRA})},
	publisher = {IEEE},
	author = {Mal, Fangchang and Karaman, Sertac},
	year = {2018},
	pages = {1--8}
}

@inproceedings{laina_deeper_2016,
	title = {Deeper depth prediction with fully convolutional residual networks},
	isbn = {1-5090-5407-3},
	booktitle = {2016 {Fourth} international conference on 3D vision (3DV)},
	publisher = {IEEE},
	author = {Laina, Iro and Rupprecht, Christian and Belagiannis, Vasileios and Tombari, Federico and Navab, Nassir},
	year = {2016},
	pages = {239--248}
}

@article{cao_estimating_2017,
	title = {Estimating depth from monocular images as classification using deep fully convolutional residual networks},
	volume = {28},
	number = {11},
	journal = {IEEE Transactions on Circuits and Systems for Video Technology},
	author = {Cao, Yuanzhouhan and Wu, Zifeng and Shen, Chunhua},
	year = {2017},
	pages = {3174--3182}
}

@inproceedings{chen_learning_2019,
	title = {Learning single-image depth from videos using quality assessment networks},
	booktitle = {Proceedings of the {IEEE} {Conference} on {Computer} {Vision} and {Pattern} {Recognition}},
	author = {Chen, Weifeng and Qian, Shengyi and Deng, Jia},
	year = {2019},
	pages = {5604--5613}
}

@inproceedings{li_megadepth:_2018,
	title = {Megadepth: {Learning} single-view depth prediction from internet photos},
	booktitle = {Proceedings of the {IEEE} {Conference} on {Computer} {Vision} and {Pattern} {Recognition}},
	author = {Li, Zhengqi and Snavely, Noah},
	year = {2018},
	pages = {2041--2050}
}

@inproceedings{xian_monocular_2018,
	title = {Monocular relative depth perception with web stereo data supervision},
	booktitle = {Proceedings of the {IEEE} {Conference} on {Computer} {Vision} and {Pattern} {Recognition}},
	author = {Xian, Ke and Shen, Chunhua and Cao, Zhiguo and Lu, Hao and Xiao, Yang and Li, Ruibo and Luo, Zhenbo},
	year = {2018},
	pages = {311--320}
}

@inproceedings{lan_position-aware_2014,
	title = {Position-{Aware} {ListMLE}: {A} {Sequential} {Learning} {Process} for {Ranking}.},
	booktitle = {{UAI}},
	author = {Lan, Yanyan and Zhu, Yadong and Guo, Jiafeng and Niu, Shuzi and Cheng, Xueqi},
	year = {2014},
	pages = {449--458}
}

@inproceedings{xia_listwise_2008,
	title = {Listwise approach to learning to rank: theory and algorithm},
	isbn = {1-60558-205-0},
	booktitle = {Proceedings of the 25th international conference on {Machine} learning},
	publisher = {ACM},
	author = {Xia, Fen and Liu, Tie-Yan and Wang, Jue and Zhang, Wensheng and Li, Hang},
	year = {2008},
	pages = {1192--1199}
}

@inproceedings{chen_ranking_2009,
	title = {Ranking measures and loss functions in learning to rank},
	booktitle = {Advances in {Neural} {Information} {Processing} {Systems}},
	author = {Chen, Wei and Liu, Tie-Yan and Lan, Yanyan and Ma, Zhi-Ming and Li, Hang},
	year = {2009},
	pages = {315--323}
}

@inproceedings{cao_learning_2007,
	title = {Learning to rank: from pairwise approach to listwise approach},
	isbn = {1-59593-793-5},
	booktitle = {Proceedings of the 24th international conference on {Machine} learning},
	publisher = {ACM},
	author = {Cao, Zhe and Qin, Tao and Liu, Tie-Yan and Tsai, Ming-Feng and Li, Hang},
	year = {2007},
	pages = {129--136}
}

\end{document}